%% file: udl_main.tex
\documentclass{article}

\usepackage{microtype}
\usepackage{graphicx}
\usepackage{subfigure}
\usepackage{booktabs} 

\input{header0.tex}

\input{header1.tex}

\renewcommand{\indicatorSymbol}{\mathbbm{1}}

\usepackage[accepted]{icml2020}

\usepackage{color}
\definecolor{darkgreen}{rgb}{0,0.6,0}

\def\Figref#1{Figure~\ref{#1}}
\def\Eqref#1{Equation~\ref{#1}}

\icmltitlerunning{Exact posterior distributions of wide Bayesian neural networks}

\begin{document}

\twocolumn[
\icmltitle{Exact posterior distributions of wide Bayesian neural networks%
}




\begin{icmlauthorlist}
\icmlauthor{Jiri Hron}{cam,goo}
\icmlauthor{Yasaman Bahri}{goo}
\icmlauthor{Roman Novak}{goo}
\icmlauthor{Jeffrey Pennington}{goo}
\icmlauthor{Jascha Sohl-Dickstein}{goo}
\end{icmlauthorlist}

\icmlaffiliation{cam}{Department of Engineering, University of Cambridge, United Kingdom}
\icmlaffiliation{goo}{Google Research, Brain Team, United States}

\icmlcorrespondingauthor{Jiri Hron}{jh2084@cam.ac.uk}

\icmlkeywords{Machine Learning, ICML}

\vskip 0.3in
]



\printAffiliationsAndNotice{}  

\begin{abstract}
Recent work has shown that the \emph{prior} over functions induced by a deep Bayesian neural network (BNN) behaves as a Gaussian process (GP) as the width of all layers becomes large.
However, many BNN applications are concerned with the BNN function space \emph{posterior}.
While some \emph{empirical} evidence of the posterior convergence was provided in the original works of \citet{neal1996} and \citet{matthews2018gaussian}, it is limited to small datasets or architectures due to the notorious difficulty of obtaining and verifying exactness of BNN posterior approximations.
We provide the missing theoretical proof that the \emph{exact} BNN posterior converges (weakly) to the one induced by the GP limit of the prior.
For empirical validation, we show how to generate \emph{exact} samples from a finite BNN on a small dataset via rejection sampling.
\end{abstract}

\vspace{-1\baselineskip}

\section{Introduction}\label{sect:intro}

A BNN is composed of a finite number of layers $\depth \in \natnum$ where the output of the $\depthSymbol$\textsuperscript{th} layer $\indexedActivation{\depthSymbol}{}{\inputSymbol}$ is function of the previous layer outputs $\indexedActivation{\depthSymbol-1}{}{\inputSymbol}$, a nonlinearity $\nonlinearity$, and parameters $\theta^\depthSymbol$. 
For a fully connected network (FCN)
\begin{align*}
    \indexedActivation{\depthSymbol}{}{\inputSymbol}
    =
    W^\depthSymbol \nonlinearity(\indexedActivation{\depthSymbol-1}{}{\inputSymbol})
    +
    b^\depthSymbol
    \, ,
    \quad 
    \depthSymbol \in [\depth + 1]
    \, ,
\end{align*}
with $[\depth + 1] = \{ 1, \ldots, \depth + 1 \}$, $\theta^\depthSymbol = \{ W^\depthSymbol, b^\depthSymbol \}$, $W^\depthSymbol \in \R{\width^\depthSymbol \times \width^{\depthSymbol - 1}}, b^\depthSymbol \in \R{\width^\depthSymbol}$, and $\nonlinearity(\indexedActivation{0}{}{\inputSymbol} ) \coloneqq \inputSymbol$ for convenience.

Where a BNN differs from its NN equivalent is in the handling of the parameters $\theta = \bigcup_{\depthSymbol} \theta^\depthSymbol$. 
In particular, a BNN treats the parameters as random variables following some prior distribution $P_\theta$, and---instead of gradient based optimisation---uses Bayes' rule to calculate the posterior $P_{\theta \given D}$ given a fixed dataset $D = \{ (\inputSymbol_i, \outputSymbol_i) \}_{i = 1}^m \subset \R{\width^0 \times \width^{\depth + 1}}$, ${\indexSet} = \{ \inputSymbol_i \}_{i=1}^m$, ${\outputSet} = \{ \outputSymbol_i \}_{i=1}^m$
\begin{align*}
    p(\theta \given D)
    =
    \frac{p(\outputSet \given \theta, \indexSet) p(\theta)}{\int \! p(\outputSet \given \theta, \indexSet) p(\theta) \, \inftsm{\theta}}
    \, ,
\end{align*}
where $p(\theta)$ and $p(\theta \given D)$ are the density functions of $P_\theta$ and $P_{\theta \given D}$, and $p(\outputSet \given \theta, \indexSet) = p(\outputSymbol_1, \ldots \outputSymbol_m \given \theta , \inputSymbol_1, \ldots, \inputSymbol_m)$ is a likelihood function appropriate for the dataset.

\section{Large width behaviour of the prior}\label{sect:prior}

Since direct interpretation of the parameter space distribution is difficult, \citet{neal1996} proposed to instead study the distribution over input-to-output mappings $P_{\activationSymbol}$, $\activationSymbol = \activationSymbol^{\depth + 1}$, \emph{induced by} computing the forward pass with the randomly sampled $\theta \sim P_\theta$, i.e., for any measurable set $B$ (the \emph{usual Borel product $\sigma$-algebra} is assumed throughout)
\begin{align*}
    P_{\activationSymbol} (B)
    =
    P_{\activationSymbol_\theta} (B)
    = 
    \int \! 
        \indicatorSymbol\{ \activationSymbol_\theta \in B \}
        \, \inftsm{P_{\theta}}(\theta) \, ,
\end{align*}
where we use $\activationSymbol_\theta = \activationSymbol$ to emphasise that $\activationSymbol$ is a fully determined by $\theta$.
Assuming a \emph{single} layer FCN architecture and an independent zero mean Gaussian prior over $\theta$ with variance of the readout weights inversely proportional to the hidden layer width $\width^{1}$, \citeauthor{neal1996} was able to show that $P_\activationSymbol$ converges weakly\footnote{A sequence of distributions $(P_n)_{n \geq 1}$ converges weakly to $P$ if $\int \! h \, \inftsm{P_n} \to \int \! h \, \inftsm{P}$ for all real-valued continuous bounded $h$.} to a centred GP distribution as $\width^{1} \to \infty$.

\citeauthor{neal1996}'s results were recently generalised to various \emph{deep} NN architectures including convolutional, pooling, residual, and attention layers \citep{matthews2018gaussian,lee2018deep,novak2019bayesian,garriga2019deep,yang2019v1,yang2019v2,hron20}.
These papers study the function space \emph{priors} $P_{\activationSymbol_\sequenceVariable}$ 
for a sequence of increasingly wide NNs,
and establish their weak convergence to a centred GP distribution with covariance determined by the architecture and the underlying sequence of parameter space \emph{priors} $P_{\theta_\sequenceVariable}$.
To ensure the asymptotic variance neither vanishes nor explodes,
$P_{\theta_\sequenceVariable}$ is assumed zero mean with marginal variances inversely proportional to layer input size.
For example, a common choice satisfying this assumption for a FCN is
\begin{align}\label{eq:gauss_init}
\begin{aligned}[c]
    W_{\sequenceVariable, ij}^\depthSymbol &\overset{\text{i.i.d.}}{\sim} \gauss \left(0 , \tfrac{\sigma_w^2}{\width_\sequenceVariable^{\depthSymbol - 1}}\right) \, , \\
    b_{i}^\depthSymbol &\overset{\text{i.i.d.}}{\sim} \gauss \left(0 , \sigma_b^2 \right) \, ,
\end{aligned}
\end{align}
for each $\depthSymbol \in [\depth + 1]$.
Throughout, we assume that the architecture and the sequence of parameter space priors $(P_{\theta_\sequenceVariable})_{\sequenceVariable \geq 1}$ was chosen such that $P_{\activationSymbol_\sequenceVariable} \Rightarrow P_{\activationSymbol}$ for some fixed distribution $P_{\activationSymbol}$, where $\Rightarrow$ denotes weak convergence.

\section{Large width behaviour of the posterior}\label{sect:posterior}

While the results establishing convergence of the function space \emph{prior} have been very influential and provided useful insights, many practical applications of BNNs require computation of expectations with respect to the function space \emph{posterior}. 
Some previous works \citep[e.g.,][]{neal1996,matthews2018gaussian} have shown good \emph{empirical} agreement of the wide BNN posterior with the one induced by $P_\activationSymbol$ for certain architectures, datasets, and likelihoods, 
but \emph{theoretical} proof of the asymptotic convergence was up until now missing.

Here we prove that the sequence of \emph{exact} function space posteriors $P_{\activationSymbol_\sequenceVariable \given D}$---induced by the sequence of \emph{exact} parameter space posteriors $P_{\theta_\sequenceVariable \given D}$---converges weakly to $P_{\activationSymbol \given D}$, the Bayesian posterior induced by the weak limit of the priors $P_{\activationSymbol_\sequenceVariable}$, under the following assumption on the likelihood.\footnote{We abuse the notation in the rest of the document by treating $\indexSet$ and $\outputSet$ as both random variables and the values these variables take. Correct interpretation should be clear from the context.}
\begin{assumption}\label{assumption:bounded_continuous}
    The targets ${\outputSet}$ depend on the network parameters $\theta_\sequenceVariable$ and the inputs $\indexSet$ only through
    \begin{align*}
        \activationSymbol_{\theta_\sequenceVariable}(\indexSet)
        = 
        \activationSymbol_\sequenceVariable(\indexSet)
        \coloneqq
        [
            \activationSymbol_{\sequenceVariable}(\inputSymbol)
        ]_{\inputSymbol \in \indexSet}
        \in \R{|D|\outputDimension}
        \, , 
    \end{align*}
    and there exists a measure $\nu$ such that the distribution of $\outputSet$ given the network outputs $P_{{\outputSet} \given \activationSymbol_\sequenceVariable(\indexSet)}$ is absolutely continuous w.r.t.\ $\nu$ for every value of $\activationSymbol_\sequenceVariable(\indexSet)$.
    Further, the resulting likelihood written as a function of $\activationSymbol_\sequenceVariable(\indexSet)$
    \begin{align*}
        \ell_\sequenceVariable (\activationSymbol_\sequenceVariable(\indexSet))
        \coloneqq
        \frac{\inftsm{P_{\outputSet \given \activationSymbol_\sequenceVariable(\indexSet)}}}{\inftsm{\nu}} (\outputSet)
        \, ,
    \end{align*}
    satisfies $\ell_\sequenceVariable = \ell$ for all $\sequenceVariable$, with $\ell \colon \R{|D|\outputDimension} \to [0, \infty)$ a \emph{continuous bounded} likelihood function.
\end{assumption}
Put another way, \Cref{assumption:bounded_continuous} says that the data is modelled as conditionally independent of $\theta_\sequenceVariable$ given $\activationSymbol_\sequenceVariable(\indexSet)$ (i.e., $\activationSymbol_\sequenceVariable(\indexSet)$ is a sufficient statistic), and the corresponding conditional distribution does not change with $\sequenceVariable$.
Fortunately, this is satisfied by many popular likelihood choices like Gaussian
\begin{align}\label{eq:likelihood}
    \ell (\activationSymbol_\sequenceVariable(\indexSet))
    \propto
    \exp \biggl\lbrace
        -\frac{1}{2\sigma^2}
        \sum_{i = 1}^{|D|} 
            \|\outputSymbol_i - \activationSymbol_\sequenceVariable(\inputSymbol_i)\|^2
    \biggr\rbrace
    \, ,
\end{align}
with $\nu$ the Lebesgue measure on $\R{|D|\outputDimension}$, 
or categorical over any number $C \in \natnum$ of categories
\begin{align*}
    \ell (\activationSymbol_\sequenceVariable(\indexSet))
    =
    \prod_{i=1}^{|D|}
    \prod_{c=1}^C
        \zeta (\activationSymbol_\sequenceVariable(x_i))_c^{\outputSymbol_{ic}}
    \, ,
\end{align*}
where $\zeta$ is the softmax function, each $\outputSymbol_i$ is assumed to be a one-hot encoding of the label, and $\nu$ is the counting measure on $[C]$.
Any \emph{continuous} transformations of network outputs (like softmax) can be assumed part of the likelihood in the statement of our main result (see \Cref{app:proofs} for the proofs).

\begin{proposition}\label{prop:posterior_convergence}
    Assume $P_{\activationSymbol_\sequenceVariable} \Rightarrow P_{\activationSymbol}$ on the usual Borel product $\sigma$-algebra,
    \Cref{assumption:bounded_continuous} holds for the chosen likelihood $\ell$, and that $\int \! \ell \, \inftsm{P_\activationSymbol} > 0$.
    Then
    \begin{align}\label{eq:posterior_convergence}
        P_{\activationSymbol_\sequenceVariable \given D} \Rightarrow P_{\activationSymbol \given D} \, ,
    \end{align}
    with $P_{\activationSymbol_\sequenceVariable \given D}$ and $P_{\activationSymbol \given D}$ the Bayesian posteriors induced by the likelihood $\ell$ and respectively the priors $P_{\activationSymbol_\sequenceVariable}$ and $P_{\activationSymbol}$.
\end{proposition}

\Cref{prop:posterior_convergence} says that whenever we can establish weak convergence of the prior, weak convergence of the posterior comes essentially for free. 
Even though we usually cannot compute the exact parameter space posterior analytically, we will often be able to compute the \emph{exact} function space posterior to which it (weakly) converges in the wide limit.

A few technical comments are due.
Firstly, we make the Borel product $\sigma$-algebra assumption only to exclude the cases where the coordinate projection $\activationSymbol \mapsto \activationSymbol(\indexSet)$ is not continuous; all the prior work cited in \Cref{sect:prior} satisfies this assumption.
Secondly, note that neither $P_\activationSymbol$ nor $P_{\activationSymbol \given D}$ need to be Gaussian; whilst $P_\activationSymbol$ often will be \citep[though exceptions exist; see][]{yang2019v1,hron20}, $P_{\activationSymbol \given D}$ will not unless the prior and the likelihood are Gaussian.
Finally, we emphasise $\activationSymbol_\sequenceVariable = \activationSymbol_{\theta_\sequenceVariable}$ by definition, i.e., even though $\theta_\sequenceVariable$ does not appear in \Cref{prop:posterior_convergence} explicitly, it is implicit in $\activationSymbol_{\sequenceVariable}$ which means we could have equally well replaced \Cref{eq:posterior_convergence} by $P_{\activationSymbol_{\theta_\sequenceVariable} \given D} \Rightarrow P_{\activationSymbol \given D}$.
In contrast to the finite case, $\activationSymbol$ in $P_{\activationSymbol \given D}$ is not to be subscripted with $\theta$ since for `$\sequenceVariable = \infty$', the mapping between an infinite parameter vector $\theta$ and $\activationSymbol$ is \emph{ill-defined},
and some at first reasonably sounding definitions entail undesirable conclusions (see \Cref{sect:other_stuff}).


While \Cref{prop:posterior_convergence} is useful, it does \emph{not} imply convergence of certain expected values such as the predictive mean and variance.
This is rectified by combining 
\Cref{corr:integral_convergence} with the results on convergence of expectations w.r.t.\ the prior (see \citep{yang2019v1,yang2019v2} for an overview).
\begin{corollary}\label{corr:integral_convergence}
    If $h$ is a real-valued \emph{continuous} function such that $\int \! | h | \, \inftsm{P_{\activationSymbol_\sequenceVariable}} \to \int \! | h | \, \inftsm{P_{\activationSymbol}} < \infty$, and the assumptions of \Cref{prop:posterior_convergence} hold, then
    \begin{align}
        \int \! h \, \inftsm{P_{\activationSymbol_\sequenceVariable \given D}}
        \to 
        \int \! h \, \inftsm{P_{\activationSymbol \given D}}
        \, .
    \end{align}
\end{corollary}

\section{Parameter space}\label{sect:other_stuff}

In light of \Cref{sect:posterior}, you may wonder about the posterior behaviour of other quantities such as the parameters $\theta_\sequenceVariable$.
Such questions are complicated by the fact that the dimension of the parameter space
grows with $\sequenceVariable$, implying that any $P_{\theta_\sequenceVariable}$ and $P_{\theta_{\sequenceVariable'}}$, $\sequenceVariable \neq \sequenceVariable'$, are not distributions on the same (measurable) space, a necessity for establishing any form of convergence.
We \emph{choose} the resolution provided by the `infinite-width, finite fan-out' interpretation \citep{matthews2018gaussian} where for each $\sequenceVariable$, a countably infinite number of hidden units (convolutional filters, attention heads, etc.) and their corresponding parameters are instantiated, but only a finite number affects the layer output.
In the FCN example, $\activationSymbol_\sequenceVariable^\depthSymbol = \{ \activationSymbol_{\sequenceVariable, i}^\depthSymbol (\inputSymbol) \}_{(\inputSymbol, i) \in \R{\width^0} \times \natnum}$ with $i \in \natnum$ the neuron index
\begin{align}\label{eq:inf_to_fin_nn}
    \activationSymbol_{\sequenceVariable, i}^{\depthSymbol + 1} (\inputSymbol)
    = 
    b_i^{\depthSymbol + 1}
    +
    \sum_{j=1}^{\width_\sequenceVariable^\depthSymbol}
        W_{\sequenceVariable, ij}^{\depthSymbol + 1} 
        \nonlinearity(\activationSymbol_{\sequenceVariable, j}^{\depthSymbol} (\inputSymbol))
    \, ,
\end{align}
where $\width_\sequenceVariable^\depthSymbol < \infty$, for \emph{all} $i \in \natnum$.

With the `infinite-width, finite fan-out' construction, each
$\theta_\sequenceVariable$
is embedded into the same
infinite dimensional space $\R{\natnum}$, and we interpret $P_{\theta_\sequenceVariable}$ 
as the corresponding sequence of prior distributions on $\R{\natnum}$ (with the usual Borel product $\sigma$-algebra).
From now on, all results should be viewed as regarding prior and posterior distributions constructed in this way unless explicitly stated otherwise.

\begin{assumption}\label{assumption:gauss_network}
    Let the assumptions of \Cref{prop:posterior_convergence} hold, and let the underlying sequence of BNNs be composed of only fully connected, convolutional, and attention layers with the number of units (neurons, filters, heads) goes to infinity with $\sequenceVariable$, or layers without trainable parameters (e.g., average pooling, residual connections).
    Further, let $P_{\theta_\sequenceVariable}$ be centred Gaussian with diagonal covariance with non-zero entries equal to $\sigma_w^2 / \width_{\sequenceVariable}^\depthSymbol$ for a fixed $\sigma_w > 0$ and appropriate $\depthSymbol$ (resp.\ diagonal of all ones under the NTK parametrisation),\footnote{In the Neural Tangent Kernel (NTK) \citep{Jacot2018ntk} parametrisation, weights are a priori i.i.d.\ $\gauss(0, 1)$, and scaled by $\sigma_w / \sqrt{\width_\sequenceVariable^\depthSymbol}$ as part of the $\theta \mapsto \activationSymbol$ mapping, ensuring the induced $P_{\activationSymbol_\sequenceVariable}$ is the same. See \citep{jsd2020infinite} for more details.}
    except for biases for which variance may be just $\sigma_b^2 \geq 0$.
\end{assumption}

\begin{proposition}\label{prop:weight_posterior}
    Let \Cref{assumption:gauss_network} hold, and denote $\tilde{\theta}_n \coloneqq \theta_n \setminus \{ b^{\depth + 1} \}$ (i.e., all the parameters except for the top-layer bias) and the corresponding marginal of $P_{\theta}$ by $P_{\tilde{\theta}}$.
    
    Then $P_{\tilde{\theta}_\sequenceVariable \given D} \Rightarrow P_{\tilde{\theta}}$ where $P_{\tilde{\theta}}$ is defined by $P_{\tilde{\theta}_\sequenceVariable} \Rightarrow P_{\tilde{\theta}}$, i.e., the parameters with prior variance inversely proportional to $\width_{\sequenceVariable}^\depthSymbol$ converge to $\delta_0$ (point mass at zero), and the others remain independent zero mean Gaussian with their prior variance (biases, and all parameters under the NTK parametrisation).
    The top-layer bias converges in distribution to the posterior induced by summing
    $\tilde{\activationSymbol}(\inputSymbol) =  \activationSymbol(\inputSymbol) - b^{\depth + 1}$, $\activationSymbol \sim P_{\activationSymbol}$, 
    with $b^{L+1}$ where the two are treated as independent under the prior, and enter the likelihood as $\ell (\tilde{\activationSymbol}(\inputSymbol) + b^{\depth + 1})$ (see the end of the proof for the details).
    $P_{\theta_\sequenceVariable}$ converges weakly to the product of the marginal limits over $\tilde{\theta}_\sequenceVariable$ and $b^{\depth + 1}$.
\end{proposition}


\definecolor{finite_c}{HTML}{377eb8}
\definecolor{nngp_c}{HTML}{f781bf}
\definecolor{ntk_c}{HTML}{4daf4a}

\begin{figure*}[ht]
    \centering
    \includegraphics[width=\textwidth]{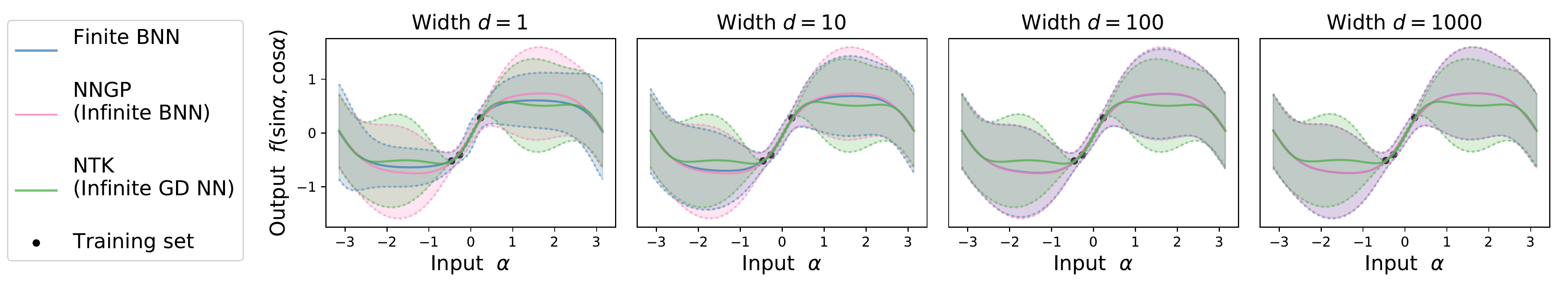}
    \vspace{-2\baselineskip}
    \caption{\textbf{Posterior sample mean and variance of a deep  {\color{finite_c}finite BNN} converge to those of an {\color{nngp_c} NNGP (infinite BNN)} as width $d$ increases} (left to right). For a given training set, posterior mean and variances are computed using rejection sampling for the {\color{finite_c}finite BNN}, and in closed form as a GP posterior (see \citep[page 16]{rasmussen2006gaussian} for the {\color{nngp_c} NNGP}, and Equation 16 in \citep{lee2019wide} for the {\color{ntk_c}NTK}). As width $d$ increases, posterior predictions of a {\color{finite_c}finite BNN} and {\color{nngp_c} NNGP} become indistinguishable, but clearly different from the {\color{ntk_c}NTK}. Presented is a fully connected network with $L=3$ hidden layers and $\phi = $ \texttt{Erf} nonlinearities. All models use a Gaussian likelihood (\Eqref{eq:likelihood}) with observation noise $\sigma^2 = 0.01$. {\color{nngp_c}NNGP} and {\color{ntk_c}NTK} predictions are computed with the Neural Tangents library \citep{neuraltangents2020}. For each width $d \in \{1, 10, 100, 1000\}$, a total of $2\cdot 10^6$ proposals are sampled, resulting in $3187$, $6796$, $8348$, and $8596$ accepts respectively. See \Figref{fig:convergence_rate} for a numerical measure of convergence.}
    \label{fig:convergence_plots}
    \vskip -0.2in
\end{figure*}

To understand \Cref{prop:weight_posterior},
we can note that the `infinite-width, finite fan-out' construction ensures the \emph{posterior marginal} over `active' parameters (those used in computation of the outputs) is exactly the posterior distribution we would have obtained had no extra parameters been introduced.
Since convergence on the infinite dimensional space typically implies convergence of all finite-dimensional marginals,\footnote{By the continuous mapping theorem for weak convergence, and, e.g., by definition for the total variation distance (modulo continuity, resp.\ measurability, of finite coordinate projections).} we can draw conclusions about the behaviour of the `active' marginals.

However, the types of conclusions we can draw also show the \emph{crucial} limitation of this approach.
For example, $P_{\theta_\sequenceVariable \given D} \Rightarrow P_{\theta}$ in the `infinite-width, finite fan-out' sense implies that for any continuous bounded real-valued function $h$, $\E h(\theta_\sequenceVariable) \to \E h(\theta)$, including $h$ which only depend on the `active' parameters of the $(\sequenceVariable')$\textsuperscript{th} network for any chosen $\sequenceVariable'$;
unfortunately, this does not guarantee the expectations are close for the $(\sequenceVariable')$\textsuperscript{th} network itself!
In other words, \Cref{prop:weight_posterior} tells us little about behaviour of \emph{finite} BNNs.

One might still wonder about the $\delta_0$ limit under the standard parametrisation, since most NN architectures output a constant if all the parameters but biases are zero.
This is due to the requirement that $P_{\theta_\sequenceVariable}$ is selected such that $P_{\activationSymbol_\sequenceVariable}$ converges, which will generally force each weight's variance to vanish as $\sequenceVariable \to \infty$ (see \Cref{eq:gauss_init} for an example).
This typically translates into the same scaling under the posterior essentially because neurons in each layer are \emph{exchangeable} \citep{matthews2018gaussian,garriga2019deep,hron20}, implying no single parameter will be `pushed too far away' from the prior by the likelihood (a similar effect can be seen in \Cref{ex:lin_reg} in \Cref{app:alternatives}).
Since concentration in an increasingly `small' region around zero is sufficient for weak convergence,\footnote{`Small' w.r.t., e.g., $d(\theta, \theta') = \sum_{i=1}^\infty 2^{-i} \min \{ 1, |\theta_i - \theta_i'| \}$ which metrises the assumed product topology.} the result follows.

Furthermore, we emphasise $P_\theta$ only describes a point in the distribution space to which the `infinite-width, finite fan-out' $P_{\theta_\sequenceVariable}$ converge, but should \emph{not} be interpreted as a distribution over parameters of an infinitely wide BNN since the $\theta \mapsto \activationSymbol$ map for `$\sequenceVariable = \infty$' is not well-defined (as mentioned in \Cref{sect:posterior}).
To see why, let us consider the single-layer FCN example with prior as in \Cref{eq:gauss_init}.
Our first impulse may be to define the $\gamma \colon \theta \mapsto \activationSymbol$ map as the pointwise limit of the functions $\gamma_\sequenceVariable \colon \theta \mapsto \activationSymbol$ where each takes in a point $\theta \in \R{\natnum}$ and computes the function implemented by the NN with corresponding index $\sequenceVariable$ (as in \Cref{eq:inf_to_fin_nn}).

If the function $\gamma(\theta)$ is to be real-valued, we need
\begin{align*}
    \gamma (\theta)(\inputSymbol)
    =
    \lim_{\sequenceVariable \to \infty}
    \gamma_\sequenceVariable(\theta)(\inputSymbol)
    =
    b^2
    +
    \lim_{\sequenceVariable \to \infty}
    \sum_{i=1}^{\width_\sequenceVariable^1}
        w_{i}^2 \nonlinearity(\activationSymbol_i^1(\inputSymbol))
\end{align*}
to be well-defined and finite which is only true if $[w_i^2 \nonlinearity(\activationSymbol_i^1(\inputSymbol))]_{i \in \natnum}$ is summable for each $\inputSymbol$ \emph{at the same time}.
This is not an issue under the standard parametrisation (since $w^2 = 0$ a.s.), but it is not satisfied under the NTK parametrisation where the support of $P_{\theta}$ is all of $\R{\natnum}$.

Since the pointwise limit approach yields $\activationSymbol = b^2$ a.s.\ when $w^2 = 0$ a.s., and $\activationSymbol = \pm \infty$ a.s.\ or is undefined (if $\lim_{n\to\infty} \gamma_\sequenceVariable(\theta)(\inputSymbol)$ does not exist) when $P_{\tilde{\theta}}$ has full $\R{\natnum}$ support (NTK parametrisation), 
we may instead try to look for $\gamma$ satisfying $\gamma(\theta) \sim P_{\activationSymbol \given D}$, $\theta \sim P_{\theta \given D}$, for all possible finite $D$, with $P_{\activationSymbol \given D}$ the limit posterior from \Cref{prop:posterior_convergence}.
As demonstrated, this is not satisfied by the pointwise limit.
It also cannot be satisfied by any other deterministic $\gamma$ if $P_{\tilde{\theta} \given D} = \delta_0$, and there will be more than one solution if $P_{\theta \given D}$ has full support on $\R{\natnum}$ (at least if we \emph{only} require agreement on a fixed countable marginal of $P_{\activationSymbol \given D}$).\footnote{Since $|D| < \infty$ and the set of \emph{finite} subsets of a countable set is countable, there is a bijection between the countable space of all $\activationSymbol$ evaluated at the countably many points for each of the countably many possible $D$, and the countable number of dimensions of $\theta$.}

All in all, we see no obvious way of defining $\gamma$ without placing restrictive assumptions on $P_{\theta \given D}$.
This is related to the dimensionality issue discussed at the beginning which forced us to adopt the additional `infinite-width, finite fan-out' assumption.
Since the `infinite-width, finite fan-out' interpretation proved much less innocuous than in the case of input-to-output mappings where it is little more than a technicality \citep{matthews2018gaussian,garriga2019deep,hron20}, we study two alternative choices in \Cref{app:alternatives}.
Unfortunately, neither yields a parameter space limit free of the pathologies we observed here.

\section{Experimental validation}

We sample from the exact finite BNN posterior using rejection sampling. 
For a given width $d_\sequenceVariable$, we use $p_{\theta_\sequenceVariable}$ as the proposal that envelops the unnormalised posterior density: 
\begin{align}
    p(\theta \given \outputSet, \indexSet) 
    &\propto  
    \ell (\activationSymbol_\theta(\indexSet)) 
    p_{\theta_\sequenceVariable}(\theta) 
    \leq 
    p_{\theta_\sequenceVariable}(\theta)
    ,
\end{align}
where $\ell (\activationSymbol_\theta(\indexSet))$ is the unnormalised Gaussian likelihood from \Eqref{eq:likelihood}.
Relatedly, prior sampling was recently used by \citet{aitchison20a} to estimate model evidence.
\Cref{fig:convergence_plots,fig:convergence_rate} confirm that as the finite BNN gets wider, its posterior mean and covariance converge to the NNGP ones.

\section{Conclusion}

We proved the sequence of exact posteriors of increasingly wide BNNs converges to the posterior induced by the infinite width limit of the prior and the same likelihood (when treated as a function of the NN outputs only).
If the computation of the infinite width limit posterior is tractable, our result opens a path to tractable function space inference even if evaluation of parameter space posterior is intractable.
We further studied conditions under which infinite width analysis in parameter space is possible and outlined several potential pitfalls.
In experiments, we have shown how to draw samples from the exact BNN posterior on small datasets, and validated our function space convergence predictions.
We hope our work provides theoretical basis for further study of exact BNN posteriors, and inspires development of more accurate BNN approximation techniques.

\definecolor{mean_c}{HTML}{e41a1c}
\definecolor{cov_c}{HTML}{984ea3}

\begin{figure}
    \centering
    \includegraphics[width=\columnwidth]{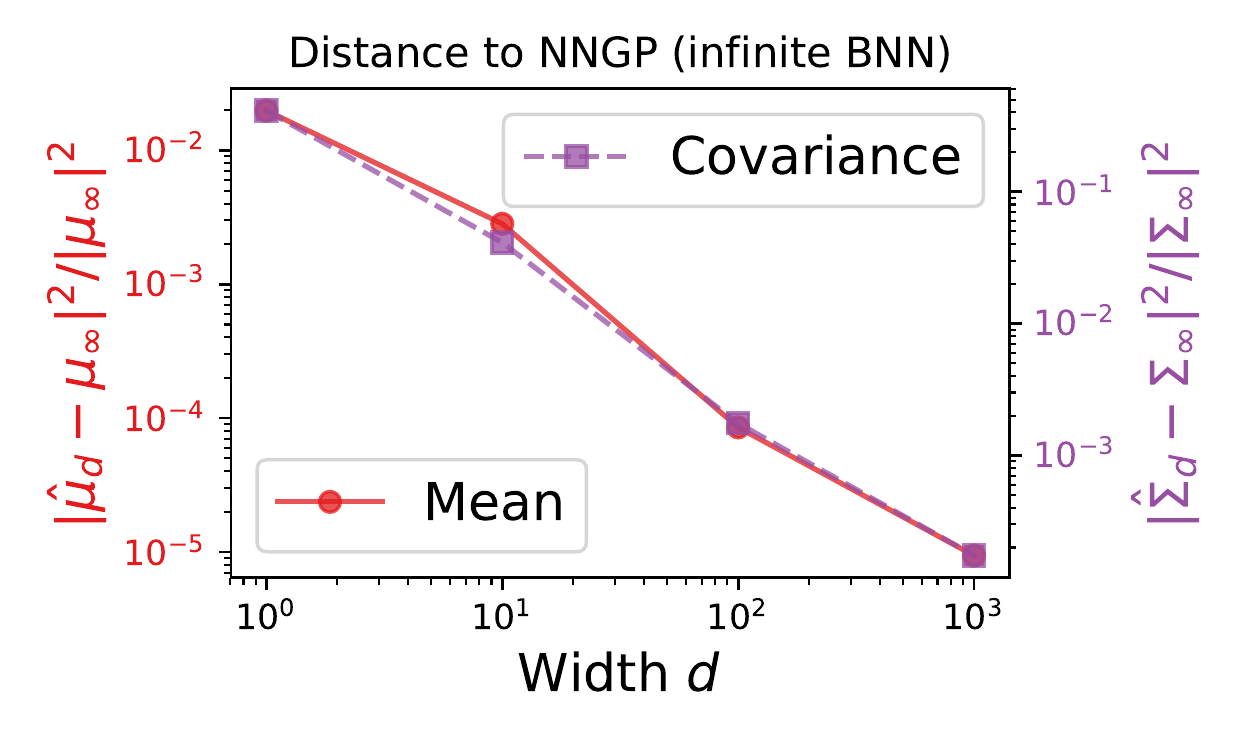}
    \caption{\textbf{Posterior sample {\color{mean_c}mean} and {\color{cov_c}covariance} of a deep finite BNN converge to those of an NNGP (infinite BNN) as width $d$ increases} (left to right, as measured by the relative Frobenius distance). Values evaluated at $100$ equidistant test points on $\left[-\pi; \pi\right]$. See \Cref{fig:convergence_plots} for a visual demonstration and further details about the setup.}
    \label{fig:convergence_rate}
\end{figure}

\section*{Acknowledgements}
    We thank Lechao Xiao for discussion, help, and feedback on the initial draft, and Wei Hu and Matej Balog for useful discussions.

\bibliography{ref}
\bibliographystyle{icml2020}

\onecolumn
\appendix

\section{Alternatives to the `infinite width, finite fan-out' interpretation}\label{app:alternatives}

The following is an (admittedly unconventional) attempt to gain intuition for the behaviour of parameter space posterior in wide BNNs by studying the simpler Bayesian linear regression model, and in particular, by measuring the discrepancy between the prior and the posterior of this model in 2-Wasserstein distance and Kullback-Leibler (KL) divergence.

\begin{example}\label{ex:lin_reg}
Let $X \in \R{m \times \sequenceVariable}$ be a matrix of $m$ inputs and $y \in \R{m}$ the vector of corresponding regression targets. 
Assume the usual Bayesian linear regression model $y \given X , w \sim \gauss(Xw , \beta I_m)$, $w \sim \gauss(0, \frac{\alpha}{\sequenceVariable}I_\sequenceVariable)$; to avoid notational clutter, we take $\alpha = \beta = 1$.
The induced posterior has closed form $w \sim \gauss(\mu_\sequenceVariable, \Sigma_\sequenceVariable)$ with
\begin{align*}
    \Sigma_\sequenceVariable 
    &= 
    \left( \sequenceVariable I_\sequenceVariable + X^\top X \right)^{-1} \\
    &=
    \tfrac{1}{\sequenceVariable}\left( I_\sequenceVariable - \tfrac{1}{\sequenceVariable}X^\top\left(I_m + \tfrac{1}{\sequenceVariable}X X^\top\right)^{-1} X \right) \, , \\
    \mu_\sequenceVariable
    &=
    \tfrac{1}{\sequenceVariable}
    X^\top \left(I_m + \tfrac{1}{\sequenceVariable} X X^\top\right)^{-1} y \, .
\end{align*}
Note that if $X$ was replaced, e.g., by the outputs of FCN's layer, $K_\sequenceVariable \coloneqq \frac{1}{\sequenceVariable} X X^\top$ would be converging almost surely to a constant $m \times m$ matrix as $\sequenceVariable \to \infty$ \citep[for an overview see, e.g.,][]{yang2019v1,yang2019v2}.
To simplify our analysis, we assume the entries of $K_\sequenceVariable$ are uniformly bounded with the implicit understanding that the results would have to be converted to high probability statements in order to hold for an actual BNN (e.g., using the results of \citeauthor{yang2019v1}).

\emph{\textbf{(I)}} We look at the \emph{squared} 2-Wasserstein distance between the posterior and the prior
\begin{align*}
    \mathcal{W}_2^2\left(\gauss(\mu_\sequenceVariable, \, \Sigma_\sequenceVariable), \gauss(0, \tfrac{1}{\sequenceVariable}I_\sequenceVariable)\right)
    &=
    \| \mu_\sequenceVariable \|_2^2
    +
    \trace\left( \tfrac{1}{\sequenceVariable} I_\sequenceVariable + \Sigma_\sequenceVariable - \tfrac{2}{\sqrt{\sequenceVariable}} \Sigma_\sequenceVariable^{1/2} \right)
    \, .
\end{align*}
By the uniform entry bound assumption, $\| \mu_\sequenceVariable \|_2 \leq \frac{1}{\sequenceVariable} [\lambda_{\text{min}} (I_m + K_\sequenceVariable)]^{-1} \| X \|_2 \| y \|_2 = \mathcal{O} (\sequenceVariable^{-1/2})$ since $\| y \|_2$ is constant and $\frac{1}{\sqrt{\sequenceVariable}} \| X \|_2 = [\lambda_{\text{max}} (K_\sequenceVariable)]^{1/2}$, where $\lambda_{\text{min}}$ and $\lambda_{\text{max}}$ are the minimum and maximum eigenvalues.
With a bit of algebra, one can also show that the value of the trace can be upper bounded by $\frac{m}{\sequenceVariable} (1 - (1 + \lambda_{\text{min}} (K_\sequenceVariable))^{-1/2} )^2 = \mathcal{O}(n^{-1})$. 
Hence the Wasserstein distance between the prior and the posterior \emph{shrinks to zero} at $\sequenceVariable^{-1/2}$ rate.\footnote{As an aside, if both $m$ and $\sequenceVariable$ were allowed to vary, the distance would be proportional to $\sqrt{m / n}$.}

If we used the `infinite-width, finite fan-out' construction, $\gauss(0, \frac{1}{\sequenceVariable} I_{\natnum})$ would be converging \emph{weakly} to $\delta_0$, and the same can be shown for the induced posterior analogously to \Cref{prop:weight_posterior}. 
On one hand, the convergence of the prior-to-posterior 2-Wasserstein distance to zero could be interpreted as a confirmation of this result.
On the other hand, $\mathcal{W}_2 (\delta_0 , \gauss(0, \frac{1}{\sequenceVariable} I_n)) = 1$ for all $\sequenceVariable$, meaning that the prior (and thus the posterior) never approaches $\delta_0$ in $\mathcal{W}_2$.
This is because $\mathcal{W}_2$ is defined w.r.t.\ the $\ell^2$ metric here which is inappropriate for $w \sim \gauss (0, \frac{1}{n} I_\natnum)$ since such $w$ is \emph{not} a.s.\ square summable.\footnote{\label{foot:wass_alt_metric}Convergence to $\delta_0$ could be recovered by using the $d(w, w') = \sum_{i=1}^\infty 2^{-i} \min \{1, |w_i - w_i'| \}$ metric induced $\mathcal{W}_2$ instead.
Since $d$ metrises the product topology w.r.t.\ which weak convergence in \Cref{prop:weight_posterior} is defined, and converegence in $\mathcal{W}_2$ is equivalent to weak convergence and convergence of the first two moments ($\E d(w, 0)^p$, $p = 1, 2$), a proof analogous to that of \Cref{prop:weight_posterior} yields the result.}

\emph{\textbf{(II)}} The KL-divergence between the posterior and the prior is
\begin{align*}
    2 \KL \left(\gauss(\mu_\sequenceVariable, \Sigma_\sequenceVariable) \, \| \, \gauss(0, \tfrac{1}{\sequenceVariable}I_\sequenceVariable)\right)
    &=
    \sequenceVariable \| \mu_\sequenceVariable \|_2^2 
    - \sequenceVariable
    + \sequenceVariable \trace(\Sigma_\sequenceVariable)
    - \sequenceVariable \log \sequenceVariable
    - \log |\Sigma_\sequenceVariable|
    \, ,
\end{align*}
where we know that the sum of all the terms from the second to the last must be non-negative (it is equal to $2\KL (\gauss(0, \Sigma_\sequenceVariable) \, \| \, \gauss(0, \tfrac{1}{\sequenceVariable}I_\sequenceVariable))$). 
Hence we can lower bound by $\sequenceVariable \| \mu_\sequenceVariable \|_2^2$ which is order one (can be obtained analogously to the upper bound derived in our discussion of  $\mathcal{W}_2$). 
This is perhaps not surprising as KL-divergence is lower bounded by (two times the square of) the total variation distance (Pinsker's inequality) in which even $\gauss (0, \frac{1}{\sequenceVariable} I_k)$---for some fixed $k \in \natnum$---does not converge to $\delta_0$.
\end{example}

While \Cref{ex:lin_reg} assumes the standard parametrisation, comparing to the conclusions that would have been drawn under the NTK parametrisation is instructive.
Since the posterior remains Gaussian (with $\mu_\sequenceVariable$ and $\Sigma_\sequenceVariable$ scaled respectively by $\sqrt{\sequenceVariable}$ and $\sequenceVariable$), it is easy to check that KL-divergence remains unchanged (as it is invariant under any injective transformation), but the 2-Wasserstein distance grows by a multiplicative factor of $\sqrt{\sequenceVariable}$ since 
\begin{align*}
    \mathcal{W}_2^2 (
        \gauss (
            \sqrt{\sequenceVariable} \mu_\sequenceVariable , \sequenceVariable \Sigma_\sequenceVariable
        )
        ,
        \gauss(0, I_\sequenceVariable)
    ) 
    &= 
    \inf_{\Gamma} \E_{(w_0, w) \sim \Gamma} \| w_0 - w \|_2^2
    \\
    &=
    \sequenceVariable \,
    \inf_{\Gamma} 
        \E_{(w_0, w) \sim \Gamma} \biggl\| 
        \frac{w_0}{\sqrt{\sequenceVariable}} 
        - 
        \frac{w}{\sqrt{\sequenceVariable}} 
    \biggr\|_2^2
    =
    \sequenceVariable \,
    \mathcal{W}_2^2 (
        \gauss (
            \mu_\sequenceVariable , \Sigma_\sequenceVariable
        )
        ,
        \gauss(0, I_\sequenceVariable)
    ) 
    \, ,
\end{align*}
where the infinum ranges over all joint distributions $\Gamma$ on $\R{2\sequenceVariable}$ which have $\gauss ( \sqrt{\sequenceVariable} \mu_\sequenceVariable , \sequenceVariable \Sigma_\sequenceVariable)$ and $\gauss(0, I_\sequenceVariable)$ as their respective $\sequenceVariable$-dimensional marginals.
In other words, the prior-to-posterior Wasserstein distance does not converge to zero when Euclidean distance is used (it will converge to zero when used with the metric from \Cref{foot:wass_alt_metric} though, which is why the above is not a contradiction of \Cref{prop:weight_posterior}; cf.\ the last statement in \textbf{(I)}, \Cref{ex:lin_reg}).

The above implies we need to be careful in interpreting rates of convergence, and in particular, that some discrepancy measures like the Wasserstein metrics necessitate choice of measurement unit for the parameters.
KL-divergence does not suffer from such issues but its relation with the total variation distance could make it excessively strict (total variation distance implies weak convergence but the reverse is not true; see the example in the last statement in \textbf{(II)} in \Cref{ex:lin_reg}).

While we cannot offer a definite answer to the above issues, it is worth pointing out that what we care about in practice is the accuracy of the \emph{function space} approximation where issues of changing dimensionality disappear, and measurement units are dictated by the dataset we are trying to model.
Hence a potentially more fruitful approach would be to refocus our attention from the parameter space to measurement and optimisation of function space approximation accuracy.

\section{Proofs}\label{app:proofs}

\begin{proof}[Proof of \Cref{prop:posterior_convergence}]
    By the definition of weak convergence, all we need to show is that for any continuous bounded function $h \colon \activationSymbol \mapsto h(\activationSymbol) \in \R{}$, the expectation converges $\int \! h \, \inftsm{P_{\activationSymbol_\sequenceVariable \given D}} \to \int \! h \, \inftsm{P_{\activationSymbol \given D}}$.
    The key observation is that \Cref{assumption:bounded_continuous} ensures each posterior $P_{\activationSymbol_\sequenceVariable \given D}$ has a density w.r.t.\ the prior \citep[e.g.][theorem~1.31]{schervish2012theory}
    \begin{align*}
        \frac{\inftsm{P_{\activationSymbol_\sequenceVariable \given D}}}{\inftsm{P_{\activationSymbol_\sequenceVariable}}} (\activationSymbol)
        =
        \frac{\ell(\activationSymbol(\indexSet))}{Z_\sequenceVariable}
        \, ,
    \end{align*}
    where $Z_\sequenceVariable \coloneqq \int \! \ell \, \inftsm{P_{\activationSymbol_\sequenceVariable}}$.
    Substituting
    \begin{align*}
        \int \! h(\activationSymbol) \, \inftsm{P_{\activationSymbol_\sequenceVariable \given D}}(\activationSymbol)
        = 
        \frac{1}{Z_\sequenceVariable}
        \int \! h(\activationSymbol) \ell(\activationSymbol(\indexSet)) \, \inftsm{P_{\activationSymbol_\sequenceVariable}}(\activationSymbol)
        \, .
    \end{align*}
    Since $\ell$ is continuous bounded and 
    $Z = \int \! \ell \, \inftsm{P_\activationSymbol} > 0$ by assumption, $Z_\sequenceVariable \to Z$ by $P_{\activationSymbol_\sequenceVariable} \Rightarrow P_{\activationSymbol}$.
    Similarly, $\activationSymbol \mapsto h(\activationSymbol) \ell(\activationSymbol(\indexSet))$ is continuous bounded, and thus also
    \begin{align*}
        \int \! h(\activationSymbol) \ell(\activationSymbol(\indexSet)) \, \inftsm{P_{\activationSymbol_\sequenceVariable}}(\activationSymbol)
        \to 
        \int \! h(\activationSymbol) \ell(\activationSymbol(\indexSet)) \, \inftsm{P_{\activationSymbol}}(\activationSymbol)
        \, .
    \end{align*}
    The proof is concluded by observing theorem 1.31 \citep{schervish2012theory} applies also to the density of $P_{\activationSymbol \given D}$ w.r.t.\ $P_\activationSymbol$.
\end{proof}

\begin{proof}[Proof of \Cref{corr:integral_convergence}]
    Following the proof strategy of \Cref{prop:posterior_convergence}
    \begin{align*}
        \int \! h \, \inftsm{P_{\activationSymbol_\sequenceVariable \given D}}
        = 
        \frac{1}{Z_\sequenceVariable}
        \int \! h({\activationSymbol}) \ell({\activationSymbol}(\indexSet)) \, \inftsm{P_{\activationSymbol_\sequenceVariable}}({\activationSymbol})
        \, ,
    \end{align*}
    we see that all we need to prove is the convergence of the integral on the right hand side ($Z_\sequenceVariable \to Z$ established in \Cref{prop:posterior_convergence}).
    Let $\activationSymbol_\sequenceVariable \sim P_{\activationSymbol_\sequenceVariable}$ and
    $\activationSymbol \sim P_{\activationSymbol}$.
    Since $h$ is continuous, $|h(\activationSymbol_{\sequenceVariable})| \Rightarrow |h(\activationSymbol)|$ by the continuous mapping theorem.
    Because the expectation of $|h|$ converges under the prior by assumption, $\{ h(\activationSymbol_\sequenceVariable) \}_{\sequenceVariable \geq 1}$ is uniformly integrable by theorem~3.6 in \citep{billingsey86}.
    Because $\ell$ is bounded by assumption, $\{ h(\activationSymbol_\sequenceVariable) \ell (\activationSymbol_\sequenceVariable(\indexSet)) \}_{\sequenceVariable \geq 1}$ is uniformly integrable as well by definition.
    Since $h(\activationSymbol_{\sequenceVariable}) \ell(\activationSymbol_\sequenceVariable(\indexSet)) \Rightarrow h(\activationSymbol) \ell(\activationSymbol(\indexSet))$ by the continuous mapping theorem again
    \begin{align*}
        \int \! h({\activationSymbol}) \ell({\activationSymbol}(\indexSet)) \, \inftsm{P_{\activationSymbol_\sequenceVariable}}({\activationSymbol})
        \to
        \int \! h({\activationSymbol}) \ell({\activationSymbol}(\indexSet)) \, \inftsm{P_{\activationSymbol}}({\activationSymbol})
        \, ,
    \end{align*}
    by theorem~3.5 in \citep{billingsey86}.
\end{proof}

\begin{proof}[Proof of \Cref{prop:weight_posterior}]
    By \citep[theorem~2.4]{billingsey86}, it will be sufficient to prove convergence of finite dimensional marginals of $P_{\theta_\sequenceVariable \given D}$.
    Denoting indices of this marginal by $J$ and the corresponding sequence of marginal distributions by $P_{\theta_\sequenceVariable^J \given D}$,
    all we need to establish is that for any continuous bounded real-valued function $h$,
    $
        \int \! h \, \inftsm{P_{\theta_\sequenceVariable^J \given D}}
        \to
        \int \! h \, \inftsm{P_{\theta^J}}
        \, .
    $
    By \Cref{assumption:bounded_continuous}, we can rewrite both the integrals in terms of the prior; for the $\int \! h \, \inftsm{P_{\theta_\sequenceVariable^J \given D}}$ this yields
    \begin{align*}
        \int \! h \, \inftsm{P_{\theta_\sequenceVariable^J \given D}}
        =
        \frac{1}{Z_\sequenceVariable}
        \int \! h (\theta^J) \ell(\activationSymbol_{\theta} (\indexSet)) \, \inftsm{P_{\theta_\sequenceVariable}} (\theta)
        \, ,
    \end{align*}
    where $\theta^J$ denotes the appropriate subset of entries of $\theta$, and $Z_\sequenceVariable = \int \! \ell (\activationSymbol_\theta(\indexSet)) \, \inftsm{P_{\theta_\sequenceVariable}} (\theta)$.
    By the same argument as in the proof of \Cref{prop:posterior_convergence}, $Z_n \to Z = \int \ell(\activationSymbol(\indexSet)) \, \inftsm{P_{\activationSymbol}} (\activationSymbol)$ where $Z > 0$ by assumption.
    Hence we can focus on 
    $$
        \int \! h (\theta^J) \ell(\activationSymbol_{\theta} (\indexSet)) \, \inftsm{P_{\theta_\sequenceVariable}} (\theta)
        =
        \int \! 
            h(\theta^J) 
            \int \! 
            \ell(\activationSymbol_\theta(\indexSet))
            \inftsm{P_{\theta_\sequenceVariable^{\natnum \setminus J}}} (\theta^{\natnum \setminus J})
            \,
        \inftsm{P_{\theta_\sequenceVariable^{J}}} (\theta^J)
    $$
    where $\theta^{\natnum \setminus J}$ are all the entries of $\theta$ not in $J$, and the equality is by boundedness of both $h$ and $\ell$, the Tonelli-Fubini theorem, and diagonal Gaussian prior (implying $P_{\theta_\sequenceVariable^{\natnum \setminus J} \given \theta_\sequenceVariable^J} = P_{\theta_\sequenceVariable^{\natnum \setminus J}}$ for all $\theta^J$).
    Also by the diagonal Gaussian assumption, we can use the change of variable formula to replace any weight $\theta_{i}$ by $\theta_i(\varepsilon) = \sigma \varepsilon_i / \sqrt{\width_\sequenceVariable^\depthSymbol}$ for an appropriate $\depthSymbol$ and $\varepsilon_i \sim \gauss(0, 1)$ i.i.d.\ (this step is of course not necessary under the NTK parametrisation).
    The r.h.s.\ above can then be rewritten as
    $$
        \int \! 
            h_n(\varepsilon^J)
            z_n(\varepsilon^J)
        \inftsm{P_{\varepsilon^{J}}} (\varepsilon^J)
        \, ,
    $$
    where
    \begin{align*}
        h_n(\varepsilon^J)
        \coloneqq 
        h(\theta^J(\varepsilon^J)) 
        \, ,
        \qquad
        z_n(\varepsilon^J) 
        \coloneqq 
        \int \! 
            \ell(\activationSymbol_{\theta^{J}(\varepsilon^J) \cup \theta^{\natnum \setminus J}}(\indexSet))
            \inftsm{P_{\theta_\sequenceVariable^{\natnum \setminus J}}} (\theta^{\natnum \setminus J})
        \, .
    \end{align*}
    Let us assume there are no top-layer biases for now, and add them back at a later point.
    Our current goal is to show that $h_\sequenceVariable (\varepsilon^J) z_n(\varepsilon^J) \to h_{\star}(\varepsilon^J) Z$ pointwise for some function $h_{\star} \colon \R{J} \to \R{}$ such that $\int \! h_* (\varepsilon^J) \, \inftsm{P_{\varepsilon^J}} (\varepsilon^J) = 
    \int \! h(\theta^J) \, \inftsm{P_\theta^J}(\theta^J)$ for $P_{\theta^J} = \delta_{0^J}$ under the standard, and $P_{\theta^J} = \gauss (0 , I_{|J|})$ under the NTK parametrisation.
    Since both $h$ and $\ell$ are bounded by assumption, $h_\sequenceVariable \cdot z_\sequenceVariable$ are uniformly bounded and thus the pointwise convergence could be combined with the dominated convergence theorem to conclude the proof.
    Since $h$ is continuous by assumption, $h_*(\varepsilon^J) = h(0)$ under the standard, and $h_*(\varepsilon^J) = h(\varepsilon^J)$ for all $\varepsilon^J$ values under the NTK parametrisation.
    One can easily verify that $\int \! h_* (\varepsilon^J) \, \inftsm{P_{\varepsilon^J}} (\varepsilon^J) = 
    \int \! h(\theta^J) \, \inftsm{P_\theta^J}(\theta^J)$ in both cases as required.
    All that remains is thus to show $z_\sequenceVariable \to Z$ pointwise.
    
    To do so, we will show that fixing a \emph{finite} set of parameters while letting the others vary does not affect the convergence of the induced input-to-output mappings $P_{\activationSymbol_\sequenceVariable \given \varepsilon^J} \Rightarrow P_{\activationSymbol}$ where $P_{\activationSymbol_\sequenceVariable \given \varepsilon^J}$ denotes the function space distribution given the fixed $\varepsilon^J$.
    We achieve this by a modification of the proof techniques in \citep{matthews2018gaussian,garriga2019deep,hron20}.
    The arguments therein are invariably build around an inductive application of the central limit theorem for infinitely exchangeable triangular arrays (eCLT) due to \citet{Blum1958} to linear projections of a finite subset of units.
    Since convergence in distribution of all such projections implies pointwise convergence of the characteristic function ($x \mapsto e^{\sqrt{-1}x}$ is continuous bounded), convergence in distribution follows.
    It will thus be sufficient to show how to modify the recursive application of the eCLT.
    
    What follows is a high-level description of this modification;
    a detailed description showcasing how to fill in the details on the FCN example can be found in \Cref{sect:z_n_convergence}.
    Let us consider layer $\depthSymbol \geq 2$ and the corresponding
    vector of activations $\activationSymbol_{\sequenceVariable}^\depthSymbol(\indexSet) = \{ \activationSymbol_{\sequenceVariable, i}^\depthSymbol(\inputSymbol) \}_{(\inputSymbol, i) \in \indexSet \times \natnum}$ (by the definition of $z_\sequenceVariable$, we only need weak convergence of $\activationSymbol_\sequenceVariable$ evaluated on the training set).
    By theorem~2.4 in \citep{billingsey86}, weak convergence is implied by weak convergence of all finite marginals.
    Denote the indices of these final marginals by $K$, and define $\activationSymbol_\sequenceVariable^{\depthSymbol, K}(\indexSet) = \{ \activationSymbol_{\sequenceVariable, i}^\depthSymbol(x) \}_{(x , i) \in \indexSet \times K}$.
    
    As mentioned, weak convergence of $\activationSymbol_\sequenceVariable^{\depthSymbol, K}(\indexSet)$ is implied by weak convergence of linear projections, so fix a vector $\alpha \in \R{K}$, and consider the scalar random variable $\langle \alpha,  \activationSymbol_\sequenceVariable^{\depthSymbol, K}(\indexSet) \rangle$.
    Conveniently, $\langle \alpha,  \activationSymbol_\sequenceVariable^{\depthSymbol, K}(\indexSet) \rangle$ can always be rewritten as
    \begin{equation*}
        \frac{1}{\sqrt{\width_\sequenceVariable^{\depthSymbol-1} }} \sum_{j \in B_n} c_{n , j} \varepsilon_j^{\depthSymbol} \, ,
    \end{equation*}
    for some (random) coefficients $c_{n, j}$, and a~subset of indices $B_n \subset \natnum$ s.t.\ $|B_n| = \width_\sequenceVariable^{\depthSymbol-1}$, $B_n \subseteq B_{n+1}$ for all $n \in \natnum$.
    Here $c_{n, j}$ are essentially a combination of the projection coefficients $\alpha$ and the inputs to the layer, whereas $\varepsilon_j^{\depthSymbol}$ are the Gaussian random variables constituting $\theta^\depthSymbol (\varepsilon^\depthSymbol)$ (either directly under NTK, or by reparametrisation under standard parametrisation); see \Cref{eq:projection,eq:summand,eq:full_rescaled_sum} for an example.
    
    Defining $B \coloneqq \bigcup_n B_n$, for all $n$ large enough
    \begin{equation}\label{eq:two_sum_decomposition}
        \frac{1}{\sqrt{\width_\sequenceVariable^{\depthSymbol-1} }} \sum_{j \in B_n} c_{n, j} \varepsilon_j^\depthSymbol 
        =
        \frac{1}{\sqrt{\width_\sequenceVariable^{\depthSymbol-1} }} \sum_{i \in J \cap B} c_{n, i} \varepsilon_i^\depthSymbol
        +
        \frac{\sqrt{\width_\sequenceVariable^{\depthSymbol-1} - |J|}}{\sqrt{\width_\sequenceVariable^{\depthSymbol - 1}}}
        \frac{1}{\sqrt{\width_\sequenceVariable^{\depthSymbol - 1} - |J|}} \sum_{j \in B_n \setminus J} c_n^j \varepsilon^j
        \, .
    \end{equation}
    Using $|J| < \infty$, the~first term on the~r.h.s.\ can be shown to converge to zero in probability.
    Since $[\width_\sequenceVariable^{\depthSymbol-1} - |J|) /  \width_\sequenceVariable^{\depthSymbol-1}]^{1/2} \to 1$, and the~remaining sum is properly scaled by $(\width_{\sequenceVariable}^{\depthSymbol - 1} - |J|)^{1/2}$,  an~argument analogous to that of \citeauthor{matthews2018gaussian} can be used to establish it converges in distribution to the~desired limit as it does not depend on any of the fixed parameters in the $\depthSymbol$\textsuperscript{th} layer, and dependence on the fixed values in the previous layers vanishes as $\sequenceVariable \to \infty$ by the recursive application of the above argument (of course there are no terms that depend on the fixed values for $j \in B_n \setminus J$ when $\depthSymbol = 2$).
    A~simple application of Slutsky's lemmas (if $X_n \Rightarrow X$ and $Y_n \to c$ in probability, $c \in \R{}$, then $X_n + Y_n \Rightarrow X + c$, and $X_n Y_n \Rightarrow c X$) then yields that for any fixed values of $\langle \alpha,  \activationSymbol_\sequenceVariable^{\depthSymbol, K}(\indexSet) \rangle$ converges in distribution to the~desired limit.
    Hence, $P_{\activationSymbol_\sequenceVariable \given \varepsilon^J} \Rightarrow P_{\activationSymbol}$ for any fixed value of $\varepsilon^J$ as desired.
        
    All that remains is to add back the top-layer biases. 
    As we have seen, the $P_{\activationSymbol_\sequenceVariable \given \varepsilon^J}$ distribution without top-layer biases converges to $P_{\activationSymbol - b^{\depth + 1}}$ (the prior limit after subtraction of top-layer biases), and thus the biases may be simply added on top.
    In the case of a Gaussian $P_{\activationSymbol}$, this will result in an additive $\sigma_b^2$ term in the covariance matrix as usual (this can be proved by standard argument via characteristic function using the assumed Gaussian diagonal prior over all parameters).
    The posterior over the top-layer biases will then be same as if we did joint posterior update over $\activationSymbol \sim P_{\activationSymbol - b^{\depth + 1}}$ and $b^{\depth + 1}$ a prior distributed according to the weak limit of the corresponding marginals of $P_{\theta_\sequenceVariable}$.
\end{proof}

\subsection{Proving $z_\sequenceVariable(\varepsilon^J) \to Z$ pointwise in a fully-connected network}\label{sect:z_n_convergence}

\textit{\textbf{Note:} All of the~references to \citep{matthews2018gaussian} here are to the~version accessible at \href{https://arxiv.org/abs/1804.11271v2}{https://arxiv.org/abs/1804.11271v2}}

The~goal of this section is to adapt the~original proof by \citet{matthews2018gaussian}.
We thus omit introduction of the~notation as well as substantial discussion of the~steps that do not require modifications.
We also modify our notation to match that of \citeauthor{matthews2018gaussian} to make comparison easier.
It is thus advisable to consult section~2 in \citep{matthews2018gaussian} which introduces the~general notation before reading on, and then referring to section~6 whenever necessary.
 
We can follow the~same steps as \citeauthor{matthews2018gaussian} right until the~application of the~Cram{\' e}r-Wold device and definition of \emph{projections} $\mathcal{T}$ and \emph{summands} $\gamma$ \citep[p.~19-20]{matthews2018gaussian}.
The~application of eCLT (resp.~its modified version \citep[p.~22, lemma~10]{matthews2018gaussian}) essentially reduces the~problem of establishment of weak convergence of $\activationSymbol_\sequenceVariable$ to that of proving of convergence of its first few finite-dimensional moments.
Following \citeauthor{matthews2018gaussian}, we define the~\emph{projections} $\projection$ and \emph{summands} $\summand$ as in their equations (25) to (27) which we restate here for convenience:
\begin{align}
    \label{eq:projection}
    \projection^{(\depthSymbol)}\atWidth 
    &\coloneqq
    \sum_{(\datapoint,\widthSymbol) \in \projectionIndeces} 
        \projectionCoefficients^{(\datapoint,\widthSymbol)}
        \left[
            \indexedActivation{\depthSymbol}{\widthSymbol}{\inputSymbol}\atWidth - \indexedBias{\depthSymbol}{\widthSymbol}
        \right]
    \, ,
    \\
    \label{eq:summand}
    \summand^{(\depthSymbol)}_{\widthSymbolB}\atWidth
    &\coloneqq
    \sum_{(\datapoint,\widthSymbol) \in \projectionIndeces} 
        \projectionCoefficients^{(\datapoint,\widthSymbol)} 
        \indexedStandardNormal{\depthSymbol}{\widthSymbol, \widthSymbolB} 
        \indexedActivity{\depthSymbol-1}{\widthSymbolB}{\inputSymbol}\atWidth
        \sqrt{\weightVarianceScaled^{(\depthSymbol)}} 
    \, ,
    \\
    \label{eq:full_rescaled_sum}
    \projection^{(\depthSymbol)}\atWidth 
    &=
    \frac{1}{\sqrt{\widthFunction^{\depthSymbol-1}(n)}}
    \sum_{\widthSymbolB=1}^{\widthFunction^{\depthSymbol-1}(n)}
        \summand^{(\depthSymbol)}_{\widthSymbolB}\atWidth
    \, .
\end{align}
Here $\activity \atWidth = \nonlinearity(\activation \atWidth)$ is the~$i$\textsuperscript{th} post-nonlinearity in $\depthSymbol$\textsuperscript{th} layer of the~$n$\textsuperscript{th} network evaluated at point $\inputSymbol$, $\widthFunction^{\depthSymbol}(n)$ is the~width of the~same layer, 
$\projectionIndeces \subset \mathcal{X} \times \natnum$ identifies the~finite marginal of the~countably infinite vector $\{ \indexedActivation{\depthSymbol}{\widthSymbol}{\inputSymbol} \}_{(\inputSymbol, \widthSymbol) \in \mathcal{X} \times \natnum}$ under consideration, and $\projectionCoefficients = \{ \projectionCoefficients^{(\datapoint,\widthSymbol)} \}_{(\datapoint,\widthSymbol) \in \projectionIndeces} \in \R{\projectionCoefficients}$ is the~Cram{\' e}r-Wold projection vector.

Note that \citeauthor{matthews2018gaussian} define $\activation \atWidth$ as the~\textbf{sum} of the~inner product of the~relevant weight vector with $\indexedActivity{\depthSymbol-1}{\widthSymbol}{\inputSymbol} \atWidth$ and the~bias term $\bias$, which is why $\bias$ is subtracted in \Cref{eq:projection}.
In contrast, we omitted subtraction of $\bias$ in the~previous section to reduce the~notational clutter.
From now on, we stick with the~notation of \Cref{eq:projection,eq:summand,eq:full_rescaled_sum}.
Last point where our notation differs from \citeauthor{matthews2018gaussian} is in omitting the~dependence of $\projection^{(\depthSymbol)}\atWidth$ and $\summand^{(\depthSymbol)}_{\widthSymbolB}\atWidth$ on $\projectionIndeces$ and $\projectionCoefficients$ (the~original notation was $\projection^{(\depthSymbol)}(\projectionIndeces,\projectionCoefficients)\atWidth$ and $\summand^{(\depthSymbol)}_{\widthSymbolB}(\projectionIndeces,\projectionCoefficients)\atWidth$).

Our goal is to prove convergence of the~outputs $\{ \indexedActivation{\largestDepthIndex + 1}{\widthSymbol}{\inputSymbol} \}_{(\inputSymbol, \widthSymbol) \in \mathcal{X} \times \natnum}$ given that a~finite subset of $\{  \indexedStandardNormal{\depthSymbol}{\widthSymbol, \widthSymbolB}  \}_{\substack{i, j \in \natnum \\ } ; 1 \leq \depthSymbol \leq \largestDepthIndex + 1}$ is fixed to an~arbitrary value.
As in \citep{matthews2018gaussian}, we approach this problem by an inductive application of their lemma~10 to the~projections $\projection$ combined with theorem~3.5 from \citep{billingsey86}.
We will thus need to prove the~sums defined in \Cref{eq:full_rescaled_sum} satisfy all the~desired properties for any choice of $\projectionIndeces$, $\projectionCoefficients \in \R{\projectionIndeces}$, and $\depthSymbol = 2, \ldots, \largestDepthIndex + 1$; recall that $\activationSymbol^{1}$ corresponds to the~pre-nonlinearities in the~first layer and thus even for a~single layer neural network, $\activationSymbol^{2}$ is the~output.
This is important because the~input dimension is fixed and thus the~distribution of $\activationSymbol^{1}$ need not be Gaussian for a~given $\varepsilon^J$ as we can trivially select $|J|$ bigger than the~input dimension and thus control value of any~finite subset of the~pre-nonlinearities in the~first layer.
As you may suspect, the~fact that we can only ever affect a~\emph{finite} subset of these activations will be crucial in the~next paragraphs.

We turn to applying lemma~10 from \citep{matthews2018gaussian} for $\depthSymbol \geq 2$.
As the~lemma applies only to exchangeable sequences, our first step will be to isolate the~non-exchangeable terms.
\citeauthor{matthews2018gaussian} prove exchangeability of the~summands $\summand_\widthSymbolB^{(\depthSymbol)} \atWidth$ over the~index $\widthSymbolB$ in their lemma~8.
The~key observation here is that the~same proof still works if we exclude all indices $j$ s.t.\ $\exists \, \widthSymbol \in \projectionIndeces_{\natnum}$ with $\indexedStandardNormal{\depthSymbol}{\widthSymbol, \widthSymbolB} \in \varepsilon^J$ (where $\projectionIndeces_\natnum$ is the~set of width indices in $\projectionIndeces$), i.e., if we exclude all summands for which at least one weight is fixed through $\varepsilon^J$.
Defining $J^{\depthSymbol} \coloneqq \{ \widthSymbolB \in \natnum \colon \exists \, \widthSymbol \in \projectionIndeces_{\natnum} \text{ s.t.\ } \indexedStandardNormal{\depthSymbol}{\widthSymbol, \widthSymbolB} \in \varepsilon^J \}$, we can rewrite \Cref{eq:full_rescaled_sum} as
\begin{equation}\label{eq:actual_decomposition}
    \projection^{(\depthSymbol)}\atWidth
    =
    \frac{
        \sum_{\widthSymbol \in J^{\depthSymbol}}
            \summand^{(\depthSymbol)}_{\widthSymbol}\atWidth
    }{\sqrt{\widthFunction^{\depthSymbol-1}(n)}}
    +
    \frac{\sqrt{\widthFunction^{\depthSymbol-1}(n) - |J^{\depthSymbol}|}}{\sqrt{\widthFunction^{\depthSymbol-1}(n)}}
    \frac{
        \sum_{\widthSymbolB \in [\widthFunction^{\depthSymbol-1}(n)] \setminus J^{\depthSymbol}}
            \summand^{(\depthSymbol)}_{\widthSymbolB}\atWidth
    }{\sqrt{\widthFunction^{\depthSymbol-1}(n) - |J^{\depthSymbol}|}}
    \, ,
\end{equation}
which mirrors the~format of \Cref{eq:two_sum_decomposition} from the~previous section.

Our next step is thus to apply Slutsky's lemmas, which in particular means that we need to show that the~first term on the~r.h.s.\ of \Cref{eq:actual_decomposition} converges in probability to zero, and the~second in distribution to the~relevant GP limit as in \citep{matthews2018gaussian}.
We will start with the~first term.
Let us define
\begin{align*}
    \summand^{(\depthSymbol)}_{j}\atWidth
    &\coloneqq
    \projectionCoefficients^\top \tilde{\activitySymbol}_j^{\depthSymbol}\atWidth
    && j \in \naturalNumbers \, ,
    \nonumber \\
    \tilde{\activitySymbol}_j^{\depthSymbol}\atWidth_i 
    &\coloneqq
    \indexedStandardNormal{\depthSymbol}{(i), j} \indexedActivity{\depthSymbol-1}{j}{\inputSymbol_{(i)}}\atWidth
    && i \in \{1, \ldots, |\projectionIndeces|\}
    \, ,
\end{align*}
as in \citep[appendix~B.1]{matthews2018gaussian} and also, to reduce notational clutter, w.l.o.g.\ assume the~weight variance scaling $\weightVarianceScaled^{(\depthSymbol)}$ is equal to one $\forall \depthSymbol$.
Now observe
\begin{equation}
    \biggl|
        \frac{1}{\sqrt{\widthFunction^{\depthSymbol-1}(n)}}
        \sum_{\widthSymbol \in J^{\depthSymbol}}
            \summand^{(\depthSymbol)}_{\widthSymbol}\atWidth
    \biggr|
    \leq
    \frac{1}{\sqrt{\widthFunction^{\depthSymbol-1}(n)}}
    \sum_{\widthSymbol \in J^{\depthSymbol}}
        \| \projectionCoefficients \|_2
        \| \tilde{\activitySymbol}_\widthSymbol^{\depthSymbol}\atWidth \|_2
    \qquad
    \text{a.s.}
    \, ,
\end{equation}
by a~simple application of the~Cauchy-Schwarz inequality.
Since $|J^{\depthSymbol}| < \infty$ and $\| \projectionCoefficients \|_2 < \infty$, an easy approach of showing that the~sum converges in probability to zero is to apply the~Markov's inequality to obtain that for any $\delta > 0$
\begin{align*}
    \Prob \biggl( 
        \sum_{\widthSymbol \in J^{\depthSymbol}}
            \| \tilde{\activitySymbol}_\widthSymbol^{\depthSymbol}\atWidth \|_2
        \geq \sqrt{\widthFunction^{\depthSymbol-1}(n)} \delta
    \biggr)
    \leq
    \frac{
        \E \biggl[ \biggl(
            \sum_{\widthSymbol \in J^{\depthSymbol}}
                    \| \tilde{\activitySymbol}_\widthSymbol^{\depthSymbol}\atWidth \|_2
        \biggr)^2 \biggr]
    }{\delta^2 \widthFunction^{\depthSymbol-1}(n)}
    \leq
    \frac{
        |J^{\depthSymbol}|
        \sum_{\widthSymbol \in J^{\depthSymbol}}
            \E \| \tilde{\activitySymbol}_\widthSymbol^{\depthSymbol}\atWidth \|_2^2
    }{\delta^2 \widthFunction^{\depthSymbol-1}(n)}
    \, ,
\end{align*}
where we have used $(\sum_{k=1}^K |x_k|)^2 \leq K (\max (|x_1|, \ldots |x_K|))^2 \leq K \sum_{k=1}^K x_k^2$.
Hence a~sufficient condition for convergence to zero in probability is that the~expected norms of the~activations over $\projectionIndeces$ converge to a~constant.
By definition
\begin{equation*}
    \E \| \tilde{\activitySymbol}_\widthSymbolB^{\depthSymbol}\atWidth \|_2^2
    =
    \sum_{(\inputSymbol, i) \in \projectionIndeces}
        \E [(\indexedStandardNormal{\depthSymbol}{i, j})^2]
        \E [(\indexedActivity{\depthSymbol-1}{j}{\inputSymbol}\atWidth)^2]
    \, .
\end{equation*}
Since $\E [(\indexedStandardNormal{\depthSymbol}{i, j})^2] \leq \max(1, \max_{j \in J} (\varepsilon^j)^2 ) < \infty$, and by the~`linear envelope condition' $\E [(\indexedActivity{\depthSymbol-1}{j}{\inputSymbol}\atWidth)^2] \leq 2 (c^2 + m^2 \E [ (\indexedActivation{\depthSymbol-1}{j}{\inputSymbol}\atWidth)^2])$, we can establish the~convergence by proving lemma~20 from \citep{matthews2018gaussian} still holds.

Since lemma~20 from \citep{matthews2018gaussian} is also necessary to prove that the~second term on the~r.h.s.\ of \Cref{eq:actual_decomposition} converges, we now turn to this latter term.
As already mentioned, our strategy for the~latter term will be to prove its convergence in distribution using lemma~10 from \citep{matthews2018gaussian}.
Aligning the~notation by substituting $X_{n, j} = \summand^{(\depthSymbol)}_{j}\atWidth$ so that
\begin{equation}\label{eq:sum_substitution_matthews}
    S_n
    =
    \frac{1}{\sqrt{\widthFunction^{\depthSymbol-1}(n) - |J^{\depthSymbol}|}}
    \sum_{\widthSymbolB \in [\widthFunction^{\depthSymbol-1}(n)] \setminus J^{\depthSymbol}}
        \summand^{(\depthSymbol)}_{\widthSymbolB}\atWidth
    \, ,
\end{equation}
we satisfy exchangeability by definition of $J^{\depthSymbol}$, and $\E \summand^{(\depthSymbol)}_{\widthSymbol}\atWidth = 0$ as well as $\E \summand^{(\depthSymbol)}_{\widthSymbolB}\atWidth \summand^{(\depthSymbol)}_{\widthSymbolB}\atWidth = 0$ hold as long as lemma~20 from \citep{matthews2018gaussian} is true so that $\E |\indexedActivity{\depthSymbol-1}{\widthSymbolB}{\inputSymbol}\atWidth|^2 < \infty$ (since $\E \indexedStandardNormal{\depthSymbol}{\widthSymbol, \widthSymbolB}  = 0$ for all $j \in \natnum \setminus J^{(\depthSymbol)})$.
Finiteness of variance and the~absolute third moments as well as $\sigma_\star^2 = \lim_n \sigma_n^2 \coloneqq \lim_{n \to \infty} \Var (\summand^{(\depthSymbol)}_{\widthSymbolB}\atWidth)$ will be established in course of proving that the~conditions b)~$\lim_{\sequenceVariable \to \infty} \E \summand^{(\depthSymbol)}_{\widthSymbol}\atWidth^2 \summand^{(\depthSymbol)}_{\widthSymbolB}\atWidth^2 = \sigma_\star^4$, and c)~$\E |\summand^{(\depthSymbol)}_{\widthSymbolB}\atWidth|^3 = \littleO(\sqrt{\widthFunction^{\depthSymbol-1}(n) - |J^{\depthSymbol}|})$ of lemma~10 from \citep{matthews2018gaussian} still hold.

We thus turn to proving conditions~b) and~c) are satisfied for any fixed value of $\varepsilon^J$.
In the~original paper, this is accomplished respectively in lemmas~15 and~16.
On closer inspection, fixing $\varepsilon^J$ can only affect the~proofs of these lemmas by invalidating lemma~20 or~21 from \citep{matthews2018gaussian}.
Lemma~20 establishes that for any fixed input $x \in \mathcal{X}$, $\E |\activation|^8$ is bounded by a~constant independent of $\sequenceVariable$ and $\widthSymbol$, for all $\depthSymbol \in [\largestDepthIndex + 1]$.
As in the~original proof, we proceed by induction.
For $\depthSymbol=1$, the~distribution of $\activation$ is Gaussian or a~Dirac's delta distribution which is either zero mean (if $\widthSymbol \notin J^{(1)}$), or centred at some point in $\R{}$ determined by the~$\indexedStandardNormal{1}{i, j} \in \varepsilon^J$.
In either case, the~eighth moments will be finite by basic properties of univariate Gaussian distributions.
This bound will be independent of $\sequenceVariable$ by definition of $\activationSymbol^{1}(x)$, and of $\widthSymbol$ by finiteness of $J^{1}$ and exchangeability of the~remaining terms.

As in the~original, we proceed by induction.
Assume that the~condition holds for all $\depthSymbol = 1, 2, \ldots, t - 1$ (for some $t \in \{2,\ldots,\depth+1\}$).
Then
\begin{equation*}
    \E |\indexedActivation{t}{\widthSymbol}{\inputSymbol}\atWidth|^8
    \leq
    2^{8 - 1} 
    \E \biggl[
    	|\biasSymbol_i^{t}|^8
    	+
    	\biggl|
    		\sum_{j=1}^{\widthFunction^{t-1}(\rowIndex)}
    			\weightSymbol_{i, j}^{t} \activitySymbol_j^{t-1}(\inputSymbol)\atWidth
    	\biggr|^8
    \biggr]
    \, .
\end{equation*}
Immediately, $\sup_i \E |\biasSymbol_i^{t}|^8 < \infty$ by $|J| < \infty$ and Gaussianity of the~other biases.
Moving on to the~second term, we can upper bound the~expectation
\begin{align*}
    \E
    	\biggl|
    		\sum_{j=1}^{\widthFunction^{t-1}(\rowIndex)}
    			\weightSymbol_{i, j}^{t} \activitySymbol_j^{t-1}(\inputSymbol)\atWidth
    	\biggr|^8
    &\leq
    2^{8-1}
    \E
    	\biggl|
    		\sum_{j \in J^{t}}
    			\weightSymbol_{i, j}^{t} \activitySymbol_j^{t-1}(\inputSymbol)\atWidth
    	\biggr|^8
    	+
    	\biggl|
    		\sum_{j \in [\widthFunction^{t-1}(\rowIndex)] \setminus J^{t}}
    			\weightSymbol_{i, j}^{t} \activitySymbol_j^{t-1}(\inputSymbol)\atWidth
    	\biggr|^8
    \, ,
\end{align*}
The~rest of the~argument in lemma~20 \citep{matthews2018gaussian} still holds for the~sum over $j \in [\widthFunction^{t-1}(\rowIndex)] \setminus J^{t}$ which will give us a~constant bound on its contribution independent of $\sequenceVariable$ and $\widthSymbol$.
For the~other term, we have
\begin{align*}
    \E
    	\biggl|
    		\sum_{j \in J^{t}}
    			\weightSymbol_{i, j}^{t} \activitySymbol_j^{t-1}(\inputSymbol)\atWidth
    	\biggr|^8
    \leq
    \frac{2^{8-1}|J^{t}|}{|\widthFunction^{t-1}(\rowIndex)|^4}
    \sum_{j \in J^{t}}
        \E [|\indexedStandardNormal{t}{i, j}|^8]
        (c^8 + m^8 \E |\indexedActivation{t-1}{\widthSymbol}{\inputSymbol}\atWidth|^8)
    \, ,
\end{align*}
and thus we can again upper bound by a~constant independent of $\sequenceVariable$ and $\widthSymbol$ using $|J^{t}| < \infty$, $\E |\indexedStandardNormal{t}{i, j}|^8 \leq \max(1, \max_{j\in J} |\varepsilon^j|^8)$, $|\widthFunction^{t-1}(\rowIndex)|^4 \geq 1$, and the~inductive hypothesis on $\E |\indexedActivation{t-1}{\widthSymbol}{\inputSymbol}\atWidth|^8$.
This concludes the~proof of lemma~20.

The~last outstanding task is thus to check that lemma~21 from \citep{matthews2018gaussian} holds for any fixed $\varepsilon^J$.
Inspecting the~original proof, the~argument therein holds when we substitute the~strengthened version of the~lemma~20 from above.
Recalling that the~updated version of lemma~20 was also the~only thing needed to finish the~proof that the~first term in \Cref{eq:sum_substitution_matthews} converges in probability to zero, the~above implies that $S_n$ from the~same equation converges in probability to the~desired limit.
We can thus proceed with application of Slutsky's lemmas as described in the~previous section, concluding $z_n \to Z$ pointwise as desired.

\end{document}

%% file: header0.tex
\usepackage{microtype}
\usepackage{graphicx}
\usepackage{booktabs} 
\usepackage{xcolor}
\usepackage{hyperref}

\usepackage[utf8]{inputenc} 
\usepackage{url}            
\usepackage{nicefrac}       

\usepackage[british]{babel}
\usepackage{natbib}
\usepackage{amsmath,amsthm,verbatim,amssymb,amsfonts,amscd}
\usepackage{thmtools,thm-restate}
\usepackage{mathtools}
\usepackage{mathrsfs}
\usepackage{commath}
\usepackage{bbm}
\usepackage{enumerate}
\usepackage{cancel}
\usepackage{cleveref} 

\allowdisplaybreaks

\newtheorem{proposition}{Proposition}

\newtheorem{corollary}{Corollary}

\newtheorem{assumption}{Assumption}
\newtheorem{example}{Example}

\DeclareMathOperator*{\E}{\mathbb{E}}
\DeclareMathOperator*{\Var}{\mathbb{V}}

\DeclareMathOperator{\Prob}{\mathbb{P}}

\DeclareMathOperator{\given}{\vert}

\DeclarePairedDelimiterX{\infdivx}[2]{(}{)}{%
	#1 \, \delimsize\| \, #2%
}
\DeclarePairedDelimiterX{\commaparen}[2]{(}{)}{%
	#1 , #2%
}
\DeclarePairedDelimiterX{\innerprod}[2]{\langle}{\rangle}{%
	#1 , #2%
}
\DeclarePairedDelimiterX{\parenththree}[3]{\lparen}{\rparen}{
	#1 ; #2 ; #3	
}

\newcommand*\xbar[1]{%
	\hbox{%
		\vbox{%
			\hrule height 0.5pt 
			\kern0.5ex
			\hbox{%
				\kern-0.1em
				\ensuremath{#1}%
				\kern-0.1em
			}%
		}%
	}%
}

\newcommand{\KL}{\mathrm{KL}}

\newcommand{\inftsm}[1]{\mathrm{d} #1}

\newcommand{\rv}[1]{\expandafter\MakeUppercase\expandafter{#1}}
\newcommand{\distr}[1]{\expandafter\MakeUppercase\expandafter{\mathrm{#1}}}

\newcommand{\trace}{\mathrm{Tr}}
\newcommand{\natnum}{\mathbb{N}}

\newcommand{\indicatorSymbol}{\mathbb{I}}

\newcommand{\R}[1]{\mathbb{R}^{#1}}

\newcommand{\gauss}{\mathcal{N}}

\newcommand\restr[2]{{
  \left.\kern-\nulldelimiterspace 
  #1 
  \right|_{#2} 
  }}


%% file: header1.tex
\usepackage{pifont}

\usepackage{enumerate}

\makeatletter
\newcommand*\bigcdot{\mathpalette\bigcdot@{.5}}
\newcommand*\bigcdot@[2]{\mathbin{\vcenter{\hbox{\scalebox{#2}{$\m@th#1\bullet$}}}}}
\makeatother

\newcommand{\activationSymbol}{f}
\newcommand{\activitySymbol}{g}
\newcommand{\weightSymbol}{w}
\newcommand{\biasSymbol}{b}

\newcommand{\widthFunction}{h}

\newcommand{\depthSymbol}{l}

\newcommand{\largestDepthIndex}{\depth}

\newcommand{\widthSymbol}{i}
\newcommand{\widthSymbolB}{j}

\newcommand{\inputSymbol}{x}

\newcommand{\outputSymbol}{y}

\newcommand{\indexedObject}[4]{{#1}^{#2}_{#3}{#4}}
\newcommand{\indexedActivation}[3]{\indexedObject{\activationSymbol}{#1}{#2}{(#3)}}
\newcommand{\indexedActivity}[3]{\indexedObject{\activitySymbol}{#1}{#2}{(#3)}}
\newcommand{\indexedStandardNormal}[2]{\indexedObject{\standardNormals}{#1}{#2}{}} 
\newcommand{\indexedBias}[2]{\indexedObject{\biasSymbol}{#1}{#2}{}} 

\newcommand{\activation}{\indexedActivation{\depthSymbol}{\widthSymbol}{\inputSymbol}}
\newcommand{\activity}{\indexedActivity{\depthSymbol}{\widthSymbol}{\inputSymbol}}

\newcommand{\bias}{\indexedBias{\depthSymbol}{\widthSymbol}}

\newcommand{\nonlinearity}{\phi}
\newcommand{\depth}{L}
\newcommand{\width}{d}

\newcommand{\outputDimension}{\width^{\depth+1}}

\newcommand{\varianceSymbol}{C}

\newcommand{\weightVarianceScaled}{\hat{\varianceSymbol}_{\weightSymbol}}

\newcommand{\sequenceVariable}{n}

\newcommand{\naturalNumbers}{\mathbb{N}}

\usepackage{thm-restate}

\newcommand{\datapoint}{x}

\newcommand{\indexSet}{\mathcal{X}}
\newcommand{\outputSet}{\mathcal{Y}}

\newcommand{\rowIndex}{n}

\newcommand{\littleO}{\mathrm{o}}

\newcommand{\standardNormals}{\varepsilon}
\newcommand{\summand}{\gamma}

\newcommand{\projection}{\mathcal{T}}
\newcommand{\projectionIndeces}{\mathcal{L}}
\newcommand{\projectionCoefficients}{\alpha}
\newcommand{\atWidth}{[\sequenceVariable]}